# Infrared Small Target Detection Based on Isotropic Constraint Under Complex Background

Fan Wang, Wei-xian Qian*

*School of Electronic and Optical Engineering, Nanjing University of Science and Technology, Nanjing 210094, China*

**Abstract**

Infrared search and tracking (IRST) system has been widely concerned and applied in the area of national defence. Small target detection under complex background is a very challenging task in the development of system algorithm. Low signal-to-clutter ratio (SCR) of target and the interference caused by irregular background clutter make it difficult to get an accurate result. In this paper, small targets are considered to have two characteristics of high contrast and isotropy, and we propose a multilayer gray difference (MGD) method constrained by isotropy. Firstly, the suspected regions are obtained through MGD, and then the eigenvalues of the original image's Hessian matrix are calculated to obtain the isotropy parameter of each region. Finally, those regions do not meet the isotropic constraint condition are suppressed. Experiments show that the proposed method is effective and superior to several common methods in terms of signal-to-clutter ratio gain (SCRG) and receiver operating characteristic (ROC) curve.

## 1. Introduction

Infrared search and tracking (IRST) system works by receiving infrared radiation from the target itself. Compared with radar system, infrared detection has many advantages including passive detection, good stealthiness, small size and resistance to electronic interference [1]. However, due to the limitation of detectors, infrared images are often with low resolution and poor quality, seriously disturbed by noise. Small targets in infrared images have low signal-to-clutter ratio (SCR), occupy few pixels and lack texture features, easy to be drowned in background clutter and noise, which leads to serious system omission and false alarm. Therefore, small target detection in complex background is always an important topic in IRST system.

The past decades have seen the rapid development of detection methods. Many classical algorithms have emerged. There are methods mainly in spatial domain, time domain and transform domain. In spatial domain, the background estimation algorithms firstly predict the background and subtract it from the original image, then the background suppression image can be obtained. Among them, Max-Median [2] improves the traditional Median filtering, taking the maximum Median of the four directions in the neighborhood as the value of the central pixel, which can effectively suppress the impulse noise. A similar method is the Max-Mean filtering [2]. Two dimensional least mean square error (TDLMS) method can adaptively adjust the weight matrix of the filter according to the input data, and thus accurately predict the flat background of the image [3]. Top-Hat is a classical morphological method, which eliminates the small structure in the image through open operation, so as to obtain the background estimation image [4]. The effect of Top-Hat is closely related to the selection of structural elements. Another kind of method in spatial domain is to directly suppress the background of the original image and enhance the target. There are methods based on contrast mechanism, such as Difference of Gaussian (DoG) [5], Laplacian of Gaussian (LoG) [6], local contrast measure (LCM) [7] and average absolute gray difference (AAGD)[10], as well as weighted image entropy based on local complexity [11]. The above algorithms are simple in principle and easy to realize, but they perform not well in the image with violent fluctuation of background gray. After transforming images from spatial domain to frequency domain, we can use various filters to get rid of the low-frequency components of the image and retain the high-frequency components, thus the flat background can be suppressed [12]. The frequency-domain filters commonly employed include ideal high-pass filter, Butterworth filter and Gauss high-pass filter. Wavelet transform is also introduced into small target detection, which can effectively suppress the rapidly changing non-stationary background [13]. But due to Heisenberg's uncertainty principle, the window function cannot be selected with both time accuracy and frequency accuracy. The above method based on transform domain has poor effect under low target SCR, and it is difficult to eliminate the high frequency components such as noise and sharp edges. Meanwhile, the computational complexity is high. The time domain methods are based on the continuity and consistency of the target motion and the target was detected through continuous multi frame images, including frame difference method [14], correlation method, etc. The combination of time domain and spatial domain is an effective means to detect continuously moving targets. For example, the track correlation method firstly obtains candidate points in a single frame image using the spatial or frequency domain method, and then matches the candidate points between different frames according to the information such as spatial position and grayscale feature, so as to track the motion trajectory of all candidate points and finally find the real target [15]. Recent years, with the popularity of deep learning, the method of training neural network with a large number of data sets to detect small targets has appeared[16]. However, there are finite features can be extracted because the small number and compact size of targets. In addition, infrared small targets often involve some sensitive items such as military and security, so it is hard to ensure that sufficient amount of data sets can be obtained in the corresponding scene. And the algorithm is hard to run in real time without the acceleration of GPU. All these problems limit the application of deep learning algorithm in the field of infrared small target detection.

Aiming at the features of small target and clutter, a novel small target detection method in a single frame is proposed in this paper. Firstly, a two dimensional Gaussian function is employed to build the gray distribution model of small target. According to the model, we propose multilayer gray difference (MGD) to enhance the salient area and suppress the flat background. Then, through calculating the Hessian matrix of the original image and its eigenvalues, the isotropy value of the region is obtained to constrain the result of MGD. Experiments demonstrate that the proposed method can effectively suppress most of the clutter and thus reduce the false alarm rate. Compared with several other widely used methods, the proposed method has better performance in signal-to-clutter ratio gain (SCRG) and receiver operating characteristic (ROC).

The main contributions of this work are summarized as follows:

(1) MGD is presented to enhance the target and extract the salient region;

(2) To get the scale of the small target, a scale estimation method is proposed;

(3) A clutter suppression method is presented based on isotropic constraint, which is calculated through Hessian matrix and its eigenvalues.

The remainder of this paper is organized as follows: In section 2, the motivation of this paper is explained. In Section 3, the proposed method is introduced. In Section 4, experiment is conducted on several real infrared sequence. At last, the research is summarized in Section 5.

## 2. Motivation

Humans can quickly find the object of interest in complex scenes thanks to the visual attention mechanism of the human visual system (HVS) [18]. This mechanism directs human eyes to focus on areas of high visual salience and optionally ignore other areas, which facilitates the extraction of critical information from redundant data. The salient region in an image refers to the area with a distinct difference in gray value from the neighborhood area, such as airplane in the clean sky, edges of buildings, etc. The presence of these objects causes a local mutation in the originally flat background grayscale surface [19]. Small targets in infrared images are one type of these salient regions. Depending on whether there is a local mutation and the source of the mutation, an image can be divided into three different areas: target area, homogeneous background and heterogeneous background. The grayscale surface of homogeneous background area is flat with no mutation, usually easy to be suppressed. The grayscale surface of heterogeneous background area fluctuates violently and the texture is complex. Most of the false alarm points are from the clutters in this region. Therefore, it is crucial to distinguish the target from the clutters in heterogeneous region accurately. Most of the clutters in background have irregular shapes and a higher gray value than the surrounding area with great gray gradient in some directions. But generally, clutters always show a correlation with surrounding regions in at least one direction. Target area shows different characteristics, which is considered to conform to the following two hypotheses:

(1) High contrast. The gray value of the small target is higher than its background neighborhood；

(2) Isotropy. Small target has gray mutation relative to the background in all directions, with similar gray distribution.

Based on the above analysis, this paper constructs a method to eliminate the regions that do not meet the two hypotheses step by step. The complete flowchart is shown in Fig. 1.

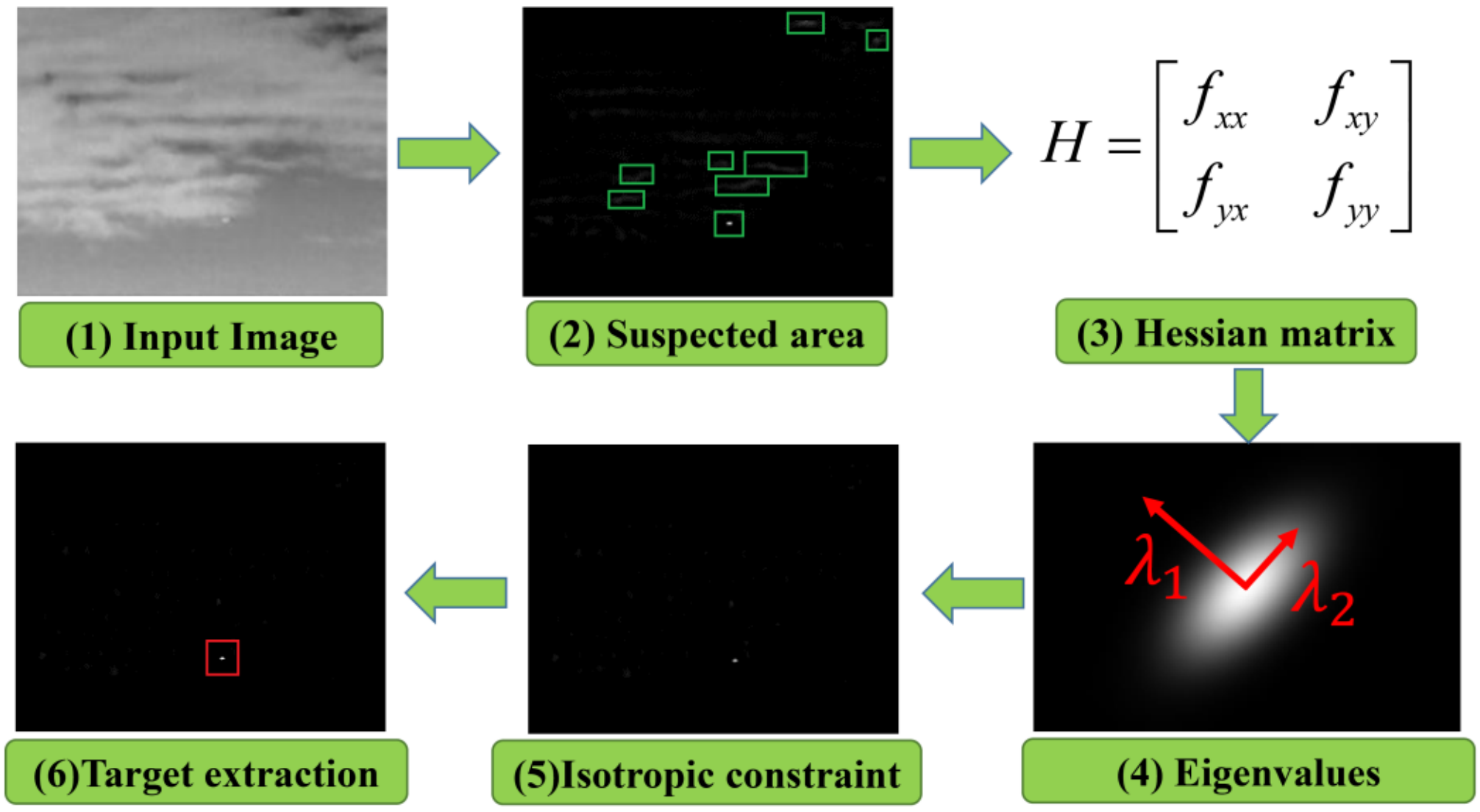

**Fig.1. Flow chart of the proposed method**

**3. Proposed method**

3.1 Multilayer gray difference

Affected by the diffraction of light and aberration of the imaging lens, the energy received by the pixels located in the small target region actually contains the infrared radiation both of the target itself and the neighboring background. Therefore, the gray value distribution of this region is similar to the two-dimensional Gaussian function [20].

The ideal small target is modeled as follows, and the model diagram is shown in Fig. 2(a) :

$$f(x,y) = Ae^{-\frac{(x-x_T)^2+(y-y_T)^2}{2\sigma^2}} + B \tag{1}$$

Where, $(x_T, y_T)$ is the coordinate of the central point of the small target area, $A$ is the amplitude of the target, $B$ is the gray mean of the background neighborhood where the target is located, and $\sigma$ is the scale parameter.

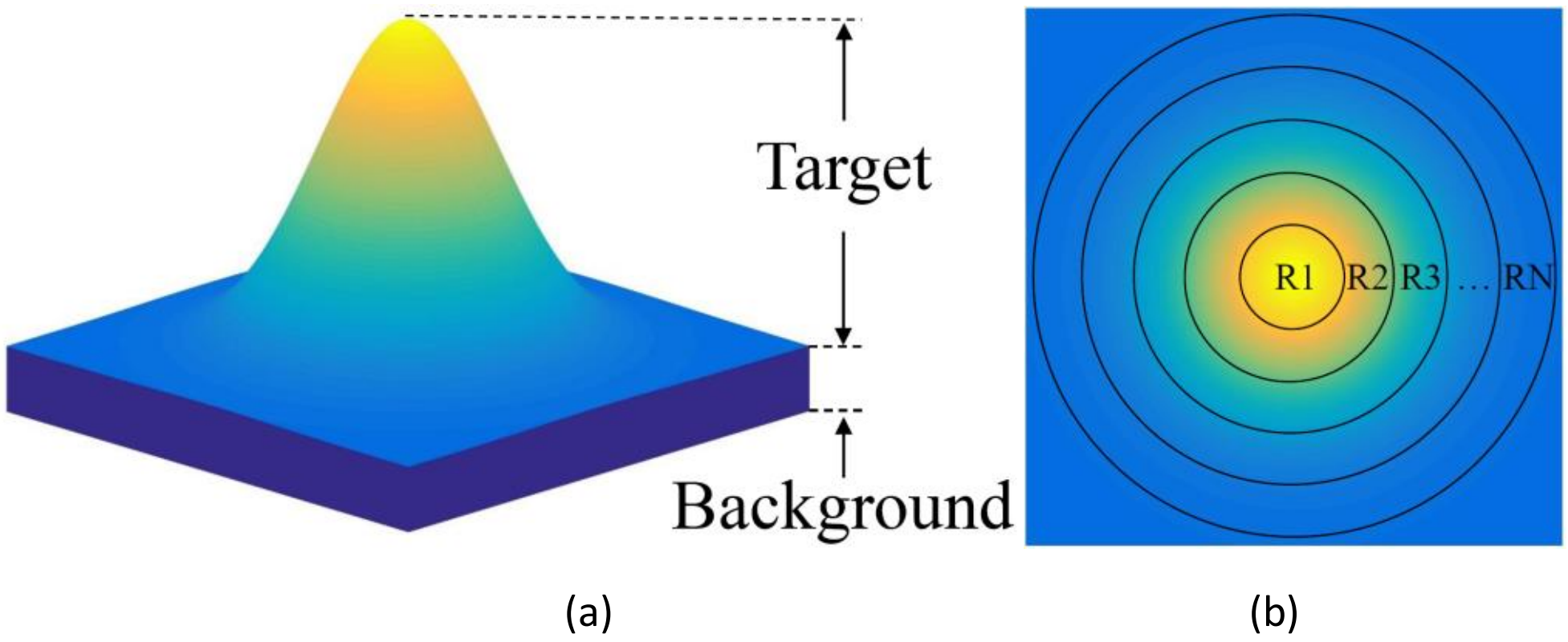

**Fig.2. (a) 3-D model of infrared small target (b) Top view of infrared small target image**

Fig. 2(b) is the top view of Fig. 2(a), which is divided into many annular regions by circles of different radius. The maximum gray value is located at the center of the target, and the grayscale at other points decays with the distance from the center. Therefore, the average grayscale of region $R_1 \sim R_N$ should meet the following conditions:

$$f_{R_1} > f_{R_2} > f_{R_3} > \ldots > f_{R_N} \tag{2}$$

$$\Rightarrow \begin{cases} f_{R_1} - f_{R_2} > 0 \\ f_{R_2} - f_{R_3} > 0 \\ \ldots \\ f_{R_{N-1}} - f_{R_N} > 0 \end{cases} \tag{3}$$

On the contrary, the grayscale of each point in homogeneous background area is similar, there are:

$$f_{R_1} \approx f_{R_2} \approx f_{R_3} \approx \ldots \approx f_{R_N} \tag{4}$$

$$\Rightarrow \begin{cases} f_{R_1} - f_{R_2} \approx 0 \\ f_{R_2} - f_{R_3} \approx 0 \\ \ldots \\ f_{R_{N-1}} - f_{R_N} \approx 0 \end{cases} \tag{5}$$

Based on the difference between the target and the homogeneous background, this paper proposes the MGD. First, a sliding window is built to traverse the whole image. Society of Photo-optical Instrumentation Engineers defines the small target to have a total spatial extent of less than 80 pixels [7]. For a $256 \times 256$ image, the small target occupies no more than 0.15% of the full image. So the sliding window is set to $9 \times 9$. All points in the sliding window are divided into different areas according to the distance from the center point, and the result is shown in Fig.3. The number in the figure represents the distance between the current pixel and the central pixel. All pixels of the same distance are divided into the same ring region marked with a unique color. In fact, each of the ring regions in Fig.3 is the boundary of the circular neighborhood with a certain radius, which can be described by the definition in literature [21]:

The neighborhood of point $(x, y)$ with radius $R$ is defined as:

$$N(R) = \{(i, j) \mid \text{round } (\sqrt{(i-x)^2 + (j-y)^2}) <= R\} (R \geq 0, R \in Z) \tag{6}$$

Where $round(x)$ is the rounding of $x$. In particular, $N(0)$ is the pixel $(x, y)$ itself.

The boundary of $N(R)$ is defined as:

$$E(R) = \{(i, j) \mid (i, j) \in N(R) - N(R-1)\} \tag{7}$$

We constructed a series of convolution kernels as shown in Fig.4 to obtain the grayscale mean $\mu_{E(R)}$ of region $E(R)$:

$$\mu_{E(1)} = f \otimes K_{r=1} \tag{8a}$$

$$\mu_{E(2)} = f \otimes K_{r=2} \tag{8b}$$

$$\ldots\ldots$$

$$\mu_{E(N)} = f \otimes K_{r=N} \qquad (8c)$$

Where the operator $\otimes$ denotes convolution. Then the MGD of point $(x, y)$ is:

$$D(x,y) = \sum_{R=1}^{N} \{ | \mu_{E(R-1)} - \mu_{E(R)} |^2 \bullet S(\mu_{E(R-1)} - \mu_{E(R)}) \} \qquad (9)$$

Where, $N$ is the number of ring regions in the sliding window, namely the maximum radius. For a $9 \times 9$ window, N is equal to 4. $S(x)$ is the step function, which is defined as:

$$S(x) = \begin{cases} 1 & x > 0 \\ 0 & x \le 0 \end{cases} \qquad (10)$$

The pseudocode of the method is given in Algorithm 1. Where, $(H, W)$ is the size of the input image. Fig.5(a) shows an infrared image, and the processing result of MGD is shown in Fig.5(b). The homogeneous background is suppressed.

| | | | | | | | | |
|---|---|---|---|---|---|---|---|---|
| 6 | 5 | 4 | 4 | 4 | 4 | 4 | 5 | 6 |
| 5 | 4 | 4 | 3 | 3 | 3 | 4 | 4 | 5 |
| 4 | 4 | 3 | 2 | 2 | 2 | 3 | 4 | 4 |
| 4 | 3 | 2 | 1 | 1 | 1 | 2 | 3 | 4 |
| 4 | 3 | 2 | 1 | 0 | 1 | 2 | 3 | 4 |
| 4 | 3 | 2 | 1 | 1 | 1 | 2 | 3 | 4 |
| 4 | 4 | 3 | 2 | 2 | 2 | 3 | 4 | 4 |
| 5 | 4 | 4 | 3 | 3 | 3 | 4 | 4 | 5 |
| 6 | 5 | 4 | 4 | 4 | 4 | 4 | 5 | 6 |

**Fig.3. Distance between each pixel and the center pixel in the sliding window**

| | | | | | | | | |
|---|---|---|---|---|---|---|---|---|
| 0 | 0 | 0 | 0 | 0 | 0 | 0 | 0 | 0 |
| 0 | 0 | 0 | 0 | 0 | 0 | 0 | 0 | 0 |
| 0 | 0 | 0 | 0 | 0 | 0 | 0 | 0 | 0 |
| 0 | 0 | 0 | $\frac{1}{8}$ | $\frac{1}{8}$ | $\frac{1}{8}$ | 0 | 0 | 0 |
| 0 | 0 | 0 | $\frac{1}{8}$ | 0 | $\frac{1}{8}$ | 0 | 0 | 0 |
| 0 | 0 | 0 | $\frac{1}{8}$ | $\frac{1}{8}$ | $\frac{1}{8}$ | 0 | 0 | 0 |
| 0 | 0 | 0 | 0 | 0 | 0 | 0 | 0 | 0 |
| 0 | 0 | 0 | 0 | 0 | 0 | 0 | 0 | 0 |
| 0 | 0 | 0 | 0 | 0 | 0 | 0 | 0 | 0 |
| 0 | 0 | 0 | 0 | 0 | 0 | 0 | 0 | 0 |
| 0 | 0 | 0 | 0 | 0 | 0 | 0 | 0 | 0 |
| 0 | 0 | 0 | $\frac{1}{12}$ | $\frac{1}{12}$ | $\frac{1}{12}$ | 0 | 0 | 0 |
| 0 | 0 | $\frac{1}{12}$ | 0 | 0 | 0 | $\frac{1}{12}$ | 0 | 0 |
| 0 | 0 | $\frac{1}{12}$ | 0 | 0 | 0 | $\frac{1}{12}$ | 0 | 0 |
| 0 | 0 | $\frac{1}{12}$ | 0 | 0 | 0 | $\frac{1}{12}$ | 0 | 0 |
| 0 | 0 | 0 | $\frac{1}{12}$ | $\frac{1}{12}$ | $\frac{1}{12}$ | 0 | 0 | 0 |
| 0 | 0 | 0 | 0 | 0 | 0 | 0 | 0 | 0 |
| 0 | 0 | 0 | 0 | 0 | 0 | 0 | 0 | 0 |
| 0 | 0 | 0 | 0 | 0 | 0 | 0 | 0 | 0 |
| 0 | 0 | 0 | $\frac{1}{16}$ | $\frac{1}{16}$ | $\frac{1}{16}$ | 0 | 0 | 0 |
| 0 | 0 | $\frac{1}{16}$ | 0 | 0 | 0 | $\frac{1}{16}$ | 0 | 0 |
| 0 | $\frac{1}{16}$ | 0 | 0 | 0 | 0 | 0 | $\frac{1}{16}$ | 0 |
| 0 | $\frac{1}{16}$ | 0 | 0 | 0 | 0 | 0 | $\frac{1}{16}$ | 0 |
| 0 | $\frac{1}{16}$ | 0 | 0 | 0 | 0 | 0 | $\frac{1}{16}$ | 0 |
| 0 | 0 | $\frac{1}{16}$ | 0 | 0 | 0 | $\frac{1}{16}$ | 0 | 0 |
| 0 | 0 | 0 | $\frac{1}{16}$ | $\frac{1}{16}$ | $\frac{1}{16}$ | 0 | 0 | 0 |
| 0 | 0 | 0 | 0 | 0 | 0 | 0 | 0 | 0 |
| 0 | 0 | $\frac{1}{32}$ | $\frac{1}{32}$ | $\frac{1}{32}$ | $\frac{1}{32}$ | $\frac{1}{32}$ | 0 | 0 |
| 0 | $\frac{1}{32}$ | $\frac{1}{32}$ | 0 | 0 | 0 | $\frac{1}{32}$ | $\frac{1}{32}$ | 0 |
| $\frac{1}{32}$ | $\frac{1}{32}$ | 0 | 0 | 0 | 0 | 0 | $\frac{1}{32}$ | $\frac{1}{32}$ |
| $\frac{1}{32}$ | 0 | 0 | 0 | 0 | 0 | 0 | 0 | $\frac{1}{32}$ |
| $\frac{1}{32}$ | 0 | 0 | 0 | 0 | 0 | 0 | 0 | $\frac{1}{32}$ |
| $\frac{1}{32}$ | 0 | 0 | 0 | 0 | 0 | 0 | 0 | $\frac{1}{32}$ |
| $\frac{1}{32}$ | $\frac{1}{32}$ | 0 | 0 | 0 | 0 | 0 | $\frac{1}{32}$ | $\frac{1}{32}$ |
| 0 | $\frac{1}{32}$ | $\frac{1}{32}$ | 0 | 0 | 0 | $\frac{1}{32}$ | $\frac{1}{32}$ | 0 |
| 0 | 0 | $\frac{1}{32}$ | $\frac{1}{32}$ | $\frac{1}{32}$ | $\frac{1}{32}$ | $\frac{1}{32}$ | 0 | 0 |

**Fig.4. Convolution kernels with different radii**

**Algorithm1 MGD**

**Input: Given frame**

**Output:** $D$

**Build a 9** $\times$ **9 sliding window.**

**for** $i = 1 : H$ **do**

  **for** $j = 1 : W$ **do**

    **for** $R = 1 : N$

      $\mu_{E(R)} = f \otimes K_{r=R}$

    **end for**

$$D(x, y) = \sum_{R=1}^{N} \{ | \mu_{E(R-1)} - \mu_{E(R)} |^2 \bullet S(\mu_{E(R-1)} - \mu_{E(R)}) \}$$

  **end for**

**end for**

**return** $D$

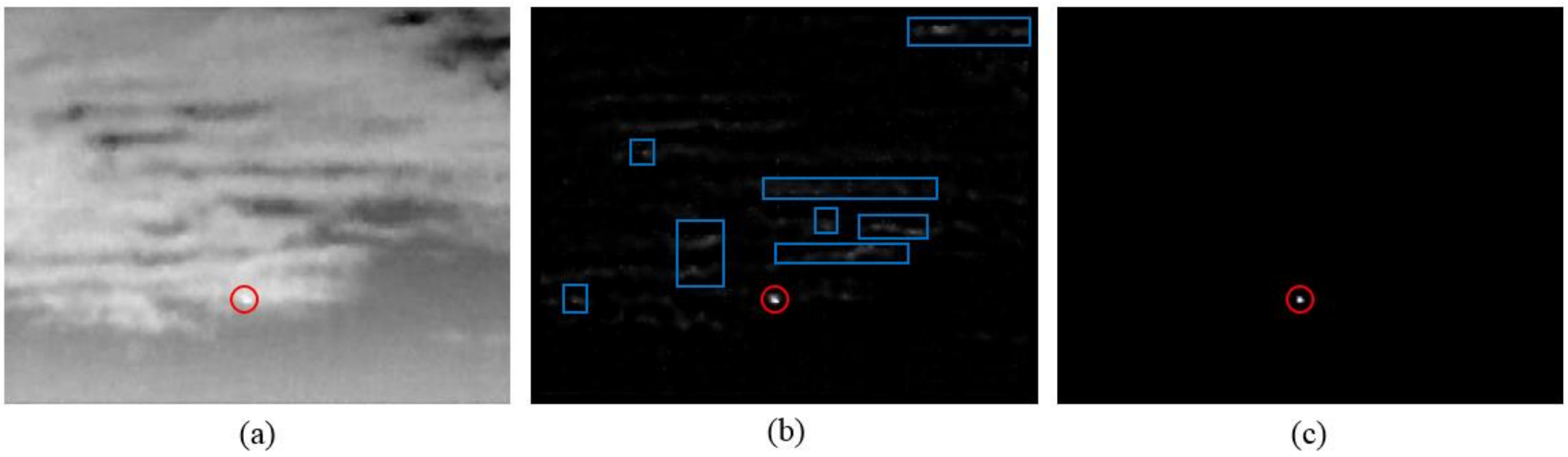

**Fig.5. (a) Original image (b) MGD image (c) Final image**

3.2 Isotropic constraint

Hessian matrix is a square matrix composed of the second partial derivatives of multivariate functions and has been widely used in image processing. Herbert Bay proposed Speed Up Robust Features (SURF) [22], extracting the points of interest in the images using Hessian matrix. Benefitting from the acceleration of integral graph in Hessian matrix calculation, this matching algorithm has high performance efficiency. In literature [23], Hessen matrix is used to design filters to detect curvilinear structures such as blood vessels in medical images and road surface cracks. There are also many other articles that use Hessian matrix to detect and analyze specific shapes in images. Zhao applied the Hessian matrix to small target detection by calculating the principal curvature of each point in infrared images [25]. These algorithms have high accuracy and robustness.

For a certain point in the image $f(x,y)$, the form of Hessen matrix $H$ and its eigenvalues $\lambda_1$ and $\lambda_2$ are shown as follows:

$$H = \begin{bmatrix} f_{xx} & f_{xy} \\ f_{yx} & f_{yy} \end{bmatrix} \tag{11}$$

$$\lambda_1 = \frac{f_{xx} + f_{yy} + \sqrt{\left(f_{xx} - f_{yy}\right)^2 + 4 f_{xy} f_{yx}}}{2} \tag{12}$$

$$\lambda_2 = \frac{f_{xx} + f_{yy} - \sqrt{\left(f_{xx} - f_{yy}\right)^2 + 4 f_{xy} f_{yx}}}{2} \tag{13}$$

Where, $f_{xx}$, $f_{yy}$ and $f_{xy}$ are the second partial derivatives of the original image.

Hessian matrix and its eigenvalues determine the structure of the gray surface in the neighborhood of the current point. The eigenvalue with the maximum absolute value and its corresponding eigenvector represent the strength and direction of the maximum curvature in the neighboring surface, while the eigenvalue with the minimum absolute value and its corresponding eigenvector represent the strength and direction of the minimum curvature. The sign of the eigenvalue represents the concavity and convexity of the gray surface. Fig.1(4) shows an elliptical bright spot with two eigenvectors at the center pointing to the directions of the maximum curvature (the gradient changes most dramatically) and the minimum curvature (the gradient changes most gently). The norm of the eigenvector represents the absolute value of the eigenvalue, which is the magnitude of the curvature as well.

After processed by MGD, there are still many irregular clutters in the background, which become the main source of false alarms. Based on the analysis at the beginning of this section, these clutters have different gray distributions in different directions, and show relevance to the background in some directions. The ideal small target satisfies isotropy. Combined with the Gaussian model of the small target, it can be concluded that the Hessian matrix and its eigenvalues in the neighborhood of target meet the following two conditions:

(1) Hessian matrix is negative definite, namely $\lambda_1 < 0 \,\&\, \lambda_2 < 0$.

(2) The absolute values of $\lambda_1$ and $\lambda_2$ are similar, namely:

$$I\left(x_T, y_T\right) = \frac{\min(|\lambda_1|,|\lambda_2|)}{\max(|\lambda_1|,|\lambda_2|)} \approx 1 \tag{14}$$

Assuming that two eigenvalues of a hessian matrix at a certain point are negative, the relationship between the region category and the value of $I$ is shown in Fig.6. Although actual small targets are not perfectly isotropic, they tend to have higher $I$ values. And most of the clutters have a lower $I$. Several typical areas in infrared image are shown in Fig.7. Their common feature is that they are brighter than the neighborhood, so they have a high response to MGD. The values of $I$ at the two small targets in Fig.7(a) and Fig.7(b) are more than 0.5, while less than 0.2 at the edge of the island in Fig.7(c) and the long and narrow building in Fig.7(d). To confirm this point further, a histogram is given in Fig.8 that reflects the distribution of $I$ of all pixels (excluding the target) of which the MGD response is higher than the mean value in an infrared image. It can be seen that pixels with $I$ above 0.8 only account for 1%. After being processed by MGD, there are still many bright salient regions retained, which are called suspected regions. We impose isotropic constraint on the suspected regions by further processing the result image of MGD, regions that do not meet the constraint are suppressed. The detailed steps are introduced below.

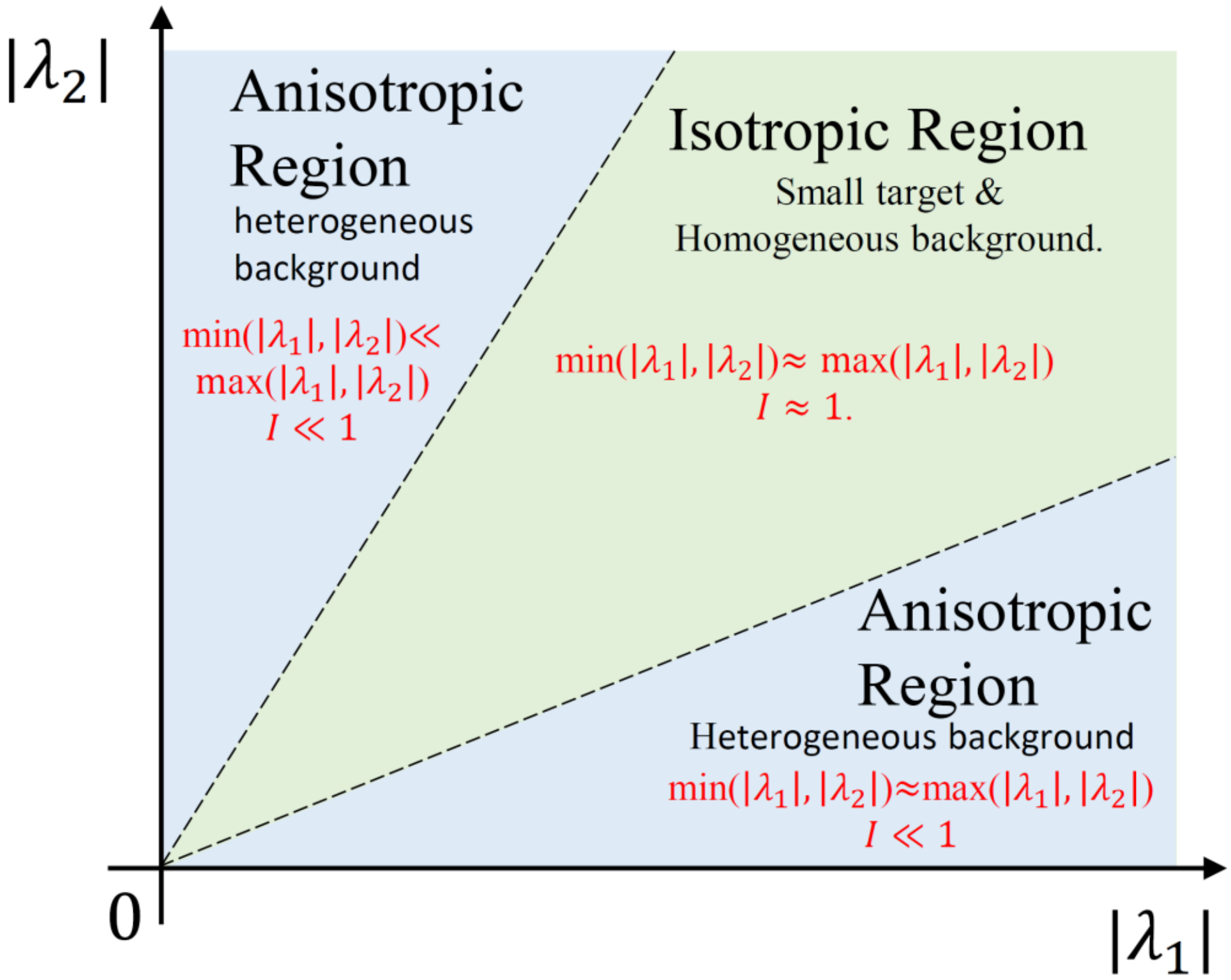

**Fig.6. The relationship between the region type and eigenvalues**

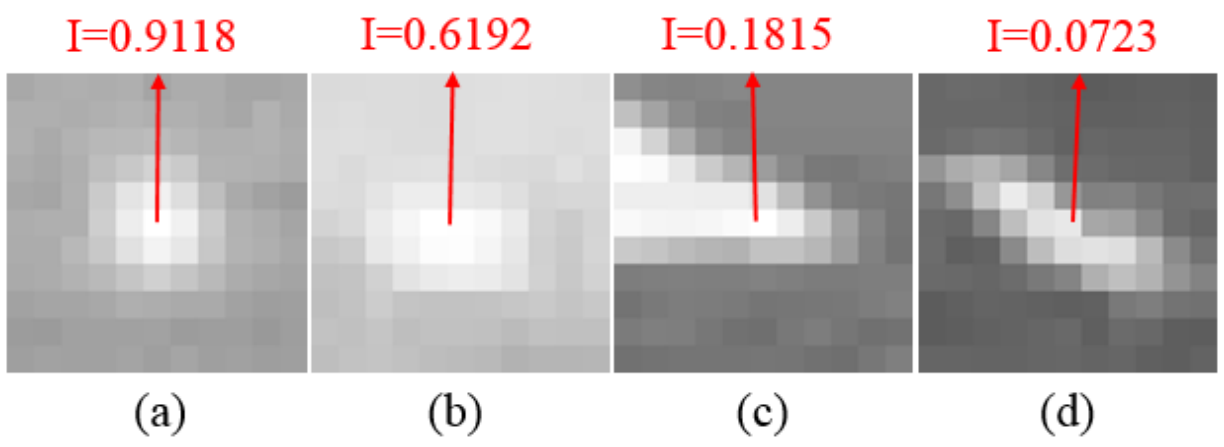

**Fig.7. $I$ of several typical regions;(a) Target 1 (b)Target 2 (c) Island (d) Building**

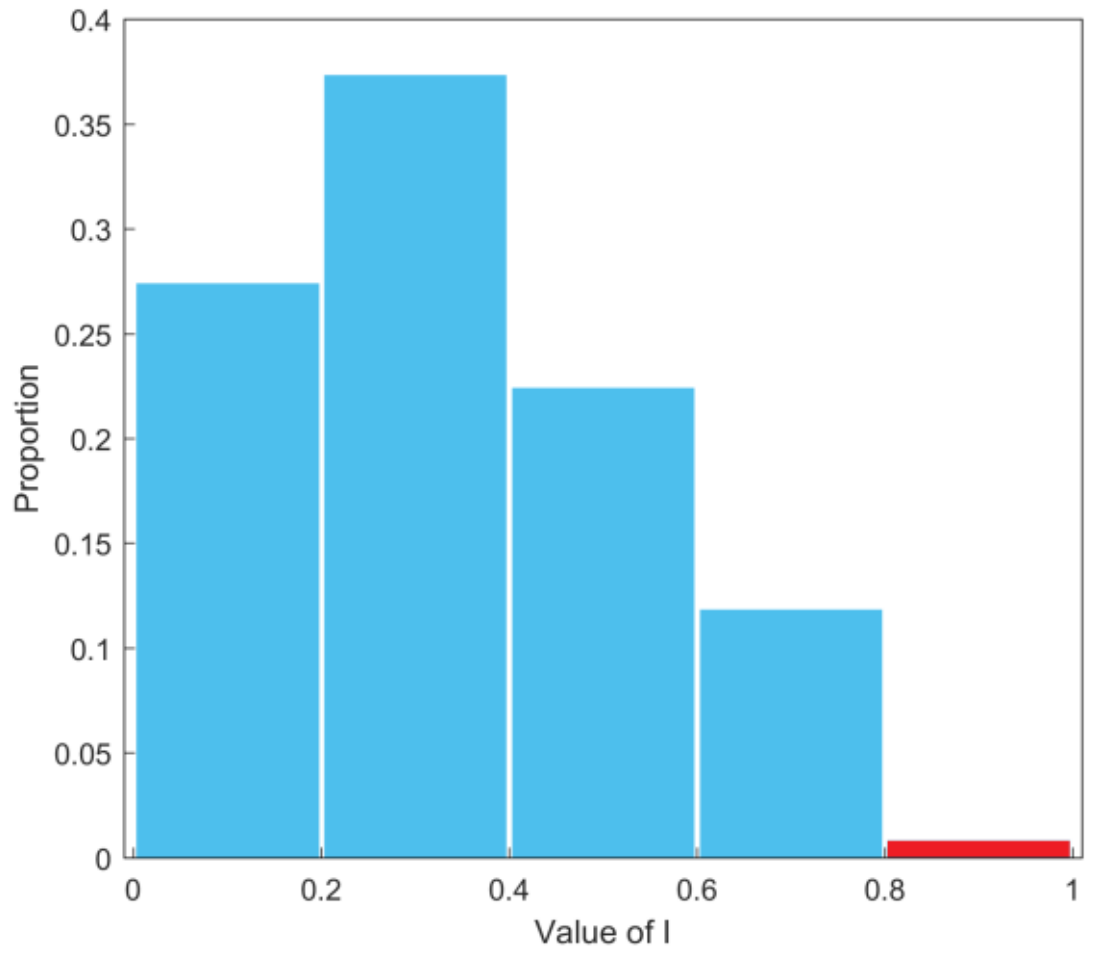

**Fig.8. Distribution of pixels in the significance region**

To obtain the Hessian matrix at a certain region, the second partial derivative of the image surface should be obtained first. According to the linear scale space theory [26], the derivative of a function is equal to the convolution of the function with the derivative of the Gaussian. The two-dimensional Gaussian function and its second partial derivatives are as follows:

$$G(x,y)=\frac{1}{2\pi\sigma^2}e^{-\frac{x^2+y^2}{2\sigma^2}} \tag{15}$$

$$G_{xx}=-\frac{1}{2\pi\sigma^4}(1-\frac{x^2}{\sigma^2})e^{-\frac{x^2+y^2}{2\sigma^2}} \tag{16}$$

$$G_{yy}=-\frac{1}{2\pi\sigma^4}(1-\frac{y^2}{\sigma^2})e^{-\frac{x^2+y^2}{2\sigma^2}} \tag{17}$$

$$G_{xy}=G_{yx}=\frac{xy}{2\pi\sigma^6}e^{-\frac{x^2+y^2}{2\sigma^2}} \tag{18}$$

Fig.9 shows the second partial derivative operator of Gaussian function generated according to equations (16) ~ (18).

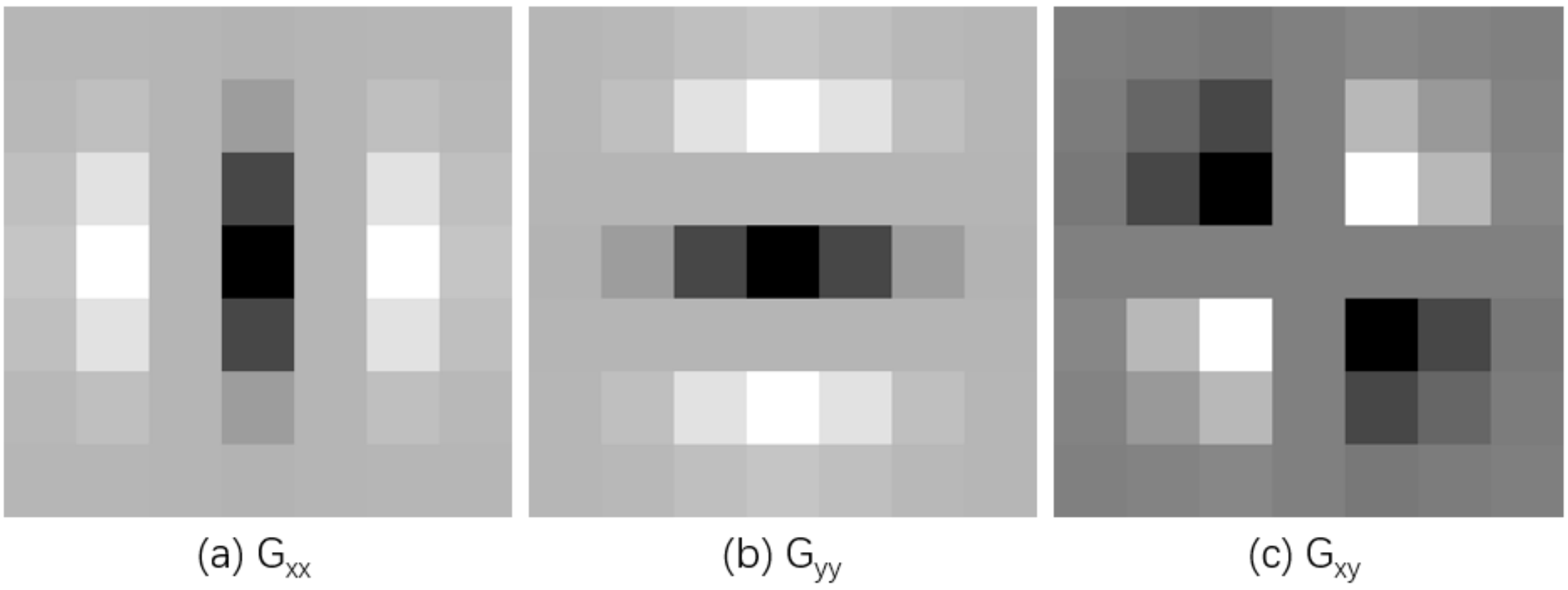

**Fig.9. Second partial derivative of Gaussian**

Then we have:

$$f_{xx}=f\otimes G_{xx} \tag{19}$$

$$f_{yy}=f\otimes G_{yy} \tag{20}$$

$$f_{xy} = f \otimes G_{xy} \tag{21}$$

After getting $f_{xx}$, $f_{yy}$ and $f_{xy}$, Hessian matrix and its eigenvalues can be obtained according to formula (11-13).

$\sigma$ in equations (15) ~ (18) is the scale parameter, the selection of which should conform to the real scale of the current local area, otherwise the result of image processing by the filter operator cannot truly reflect the feature of the current region. For example, for a small target of less than $3 \times 3$, if the $\sigma$ is too large, the partial derivative calculated actually reflects the gray features of the background in which the target is located, which may lead to the target being mistakenly suppressed. Therefore, for each point in the suspected region, the scale parameter $\sigma$ should be selected independently. This paper presents a scale estimation method.

Taking the small target center $(x_T, y_T)$ as the origin of coordinates, the rectangular coordinate system in formula (1) is transformed into the polar coordinate system. Make:

$$x - x_T = r\cos\theta \tag{22}$$

$$y - y_T = r\sin\theta \tag{23}$$

Then the converted form is:

$$\rho(r) = Ae^{-\frac{r^2}{2\sigma^2}} + B \tag{24}$$

After subtracting background grayscale $B$, the ratio $P(r)$ of amplitude at radius $r$ to amplitude at target center is:

$$P(r) = \frac{\rho(r) - B}{\rho(0) - B} = \frac{Ae^{-\frac{r^2}{2\sigma^2}}}{A} = e^{-\frac{r^2}{2\sigma^2}} \tag{25}$$

$$\Rightarrow \sigma = \frac{r}{\sqrt{-2\ln P(r)}} \tag{26}$$

Formula (26) indicates that the scale parameter $\sigma$ of the target can be obtained from $P(r)$. For an ideal Gaussian target, the resulting $\sigma$ does not change with $r$. But the real target inevitably deviates from the ideal small target model to some extent, thus different $r$ produce different $\sigma$. In this case, the multi-scale estimation method is adopted in this paper. Since the digital image is a discrete function, the radius $r$ only can be taken as a non-negative integer. The gray mean of all pixels in $E(r)$ is taken as $\rho(r)$, namely:

$$\rho(r) = \mu_{E(r)} \tag{27}$$

The value of the above formula can be obtained by the convolution of the kernel shown in Fig.4 and the original image. The kernel with the maximum radius is used to calculate the grayscale of the background. That is:

$$\rho(1) = f \otimes K_{r=1} \tag{28a}$$

$$\rho(2) = f \otimes K_{r=2} \tag{28b}$$

$$\ldots\ldots$$

$$\rho(n) = f \otimes K_{r=n} \tag{28c}$$

$$B = f \otimes K_{r=n+1} \tag{28d}$$

Substituting $\rho(1) \sim \rho(n)$ and $B$ into equations (25) and (26), we can get $\sigma(1) \sim \sigma(n)$. And the final

scale parameter is:

$$\sigma = \min\{\sigma(1), \sigma(2), ......, \sigma(n)\} \tag{29}$$

The process of computing $\sigma$ is explained in Algorithm 2.

**Algorithm 2** Scale estimation

**Input: Given patch**

**Output:** $\sigma$

$B = f \otimes K_{r=n+1}$

**for** $r = 1 : N-1$

$\rho(r) = f \otimes K_{r=n}$

$P(r) = \dfrac{\rho(r) - B}{\rho(0) - B}$

$\sigma(r) = \dfrac{r}{\sqrt{-2\ln P(r)}}$

**end for**

$\sigma = \min\{\sigma(1), \sigma(2), ......, \sigma(N-1)\}$

**return** $\sigma$

The corresponding scale parameter $\sigma$ is obtained at each point in the suspected region using the above method. The hessian matrix and its eigenvalues can be calculated by generating the second partial derivative operators of two-dimensional Gaussian functions according to the formula (16) ~ (18). Then we get the isotropic evaluation parameter $I$. The final form of the proposed method is:

$$D^{'}(x, y) = \begin{cases} 0 & D(x, y) < TH_A \\ D(x, y)I(x, y)S(-\lambda_1)S(-\lambda_2) & D(x, y) > TH_A \end{cases} \tag{30}$$

Where, $TH_A$ is the threshold to identify the salient area, which can be obtained through Otsu's method [27].

The complete computation is described in Algorithm 3. The processing effect is shown in Fig.5(c). Many irregularly shaped clutters in the original image are suppressed.

Finally, threshold segmentation and connected domain extraction are carried out, and the center of mass of the connected domain is the location of the detected target.

**Algorithm 3** Isotropic constraint

**Input: Given frame**

**Output:** $D'$

Compute $D$ according to Algorithm 1.

**for** $i = 1 : H$ **do**

**for** $j = 1 : W$ **do**

**if** $D(i, j) > TH_A$ **then**

Compute $\sigma$ according to Algorithm 2.

Get $G_{xx}$, $G_{xy}$ and $G_{yy}$ according to formula (16-18).

Get $f_{xx}$, $f_{xy}$ and $f_{yy}$ according to formula (19-21).

Compute the eigenvalues $\lambda_1$ and $\lambda_2$ of Hessian matrix according to formula (11-13).

$$I(i,j)=\frac{\min(|\lambda_1|,|\lambda_2|)}{\max(|\lambda_1|,|\lambda_2|)}$$

$$D^{'}(i,j)=\begin{cases}0 & D(i,j)<TH_A\\ D(i,j)I(i,j)S(-\lambda_1)S(-\lambda_2) & D(i,j)>TH_A\end{cases}$$

**else**

$$D'(i,j)=0$$

**end if**

**end for**

**end for**

**return** $D'$

3.3 Analysis of detection principle

The method proposed in this paper distinguishes the small target from the background by screening the original image in two steps using MGD and isotropic constraint. In this section, we discuss the effect of the proposed method when the pixel $(x,y)$ is located in different regions.

1) Pixel $(x,y)$ is located in the small target region. This region satisfies both high contrast and isotropy conditions, so it has a large value in the final output image.

2) Pixel $(x,y)$ is located in the homogeneous background. The gray surface in the sliding window is relatively flat. Therefore, the multilayer gray difference is close to 0 and is suppressed.

3) Pixel $(x,y)$ is located in the heterogeneous background. There are a lot of strong clutters in this region. The grayscale of the center of the sliding window may be higher than that of the surrounding pixels, so it is retained in the MGD map. However, most of the clutters are generally irregular in shape and the gray distribution does not meet the isotropic condition, so it will be suppressed in the final image.

According to Fig.8, there is still a very small portion of background clutters with both high contrast and isotropy. The appearance of this type of clutters is very close to the real small target, and it is difficult to distinguish it by using the spatial or frequency domain method within a single frame. One solution is to introduce time domain information and further screen the detection results of a single frame according to the difference of motion features between target and background. This mean can greatly improve the reliability of detection results. Our method can minimize the number of false alarms in the single frame detection phase to improve the overall robustness of the system algorithm.

**4. Experiments**

To verify the effectiveness and robustness of the proposed method, experiments are carried out. The image sequences used in the experiments are introduced in section 4.1. The definition of the employed evaluation metrics is given in section 4.2. Section 4.3 shows the detection effect of the proposed method on different sequences and the comparison with other commonly used algorithms. The experiments are conducted on a computer with Intel Core I7-6700, 3.40GHz processor and 16GB memory. The code was implemented on MATLAB R2016a.

### 4.1 Image sequence

The first line of Fig.10 shows the six real infrared image sequences adopted in this paper, all containing moving small targets. Sequence 1 and 2 represent the cloudy sky background. The cloud edges with drastic fluctuation of gray value and irregular block clouds are the main sources of false alarms in the sky scene. Sequence 3, 4, and 6 represent the ground background. Exposed rocks, artificial structures and roads in the image are easy to generate false alarm points. Sequence 5 represents the sea-sky background, in which the edge of the island and the ripples and glints of the sea surface are common clutters. The above sequences respectively represent different scenes and contain multiple categories of clutters, which can adequately verify the detection capability of the proposed method. The information of the experimental sequences is given in Table 1.

Table 1 The information of the test sequence

| Image sequence | Image size | Frame numbers | Target numbers | Description of the clutter |
|---|---|---|---|---|
| Seq 1 | $256\times 200$ | 30 | 1 | Cloud |
| Seq 2 | $320\times 256$ | 100 | 1 | Cloud |
| Seq 3 | $256\times 256$ | 100 | 1 | Ground objects |
| Seq 4 | $320\times 256$ | 281 | 1 | Man-made structures |
| Seq 5 | $400\times 400$ | 100 | 1 | Cloud and island |
| Seq 6 | $256\times 256$ | 200 | 1 | Ground objects |

### 4.2 Evaluation metric

In this paper, the signal to clutter ratio gain (SCRG) and the receiver operating characteristic (ROC) curve were used to evaluate the algorithm[28].

SCR and SCRG are defined as follows:

$$SCR=\frac{|I_t-\mu_B|}{\sigma_B} \quad (31)$$

$$SCRG=\frac{SCR_{out}}{SCR_{in}} \quad (32)$$

Where, $I_t$ is the maximum gray value of the target area, $\mu_B$ and $\sigma_B$ are the gray mean and standard deviation of the background neighborhood (excluding the target area). Background suppression and target enhancement algorithms can enhance the target and thus improve the target's SCR. The higher SCRG indicates the better performance of the method.

The ROC curve describes the relationship between detection rate $P_d$ and false alarm rate $P_f$ . A method with good performance has higher detection rate under the same false alarm rate, thus having a higher ROC curve. $P_d$ and $P_f$ are defined as :

$$P_d = \frac{N_d}{N_t} \tag{33}$$

$$P_f = \frac{N_f}{N_i} \tag{34}$$

Where, $N_d$ is the number of successfully detected targets (connected domain), $N_t$ is the actual number of targets (connected domain) in the current frame, $N_f$ is the number of pixels contained in all false alarm points, and $N_i$ is the number of pixels in the image. In the small target detection task, the target appears as a bright spot due to the small size and lack of contour features. So the target can be simplified to its centroid point. In the result image of threshold segmentation, when the centroid of the connected domain is located in the $3 \times 3$ neighborhood of the true location of the target, the target is considered to be successfully detected.

4.3 Comparison with other methods

To illustrate the performance of the proposed method, five commonly used detection methods are selected for quantitative and qualitative comparison, including Top-Hat, multi-scale AAGD, Max-Median, LCM, and DOG.

Qualitative analysis sis the visual effect of the of the algorithm processing result, as shown in line 2 to 6 of Figure 10. All the targets in these images are enhanced to some extent. Considering the background, a large number of clutter is retained in the Top-Hat and LCM images. Multi-scale AAGD performs well in general in the first 5 sequences, but it cannot suppress the clutter with strong edges. In particular, the response of some block and stripe clutter in sequence 6 significantly exceeds the target region. Max-median and DoG contain a subtraction step, which results in negative pixels. Hence most areas of the image are gray, and the visual effect is not obvious. But the texture of the background is still visible. The processing effect of the proposed method is the best for all sequences. The target is enhanced while the background area is clean, which means most of the clutters are successfully suppressed.

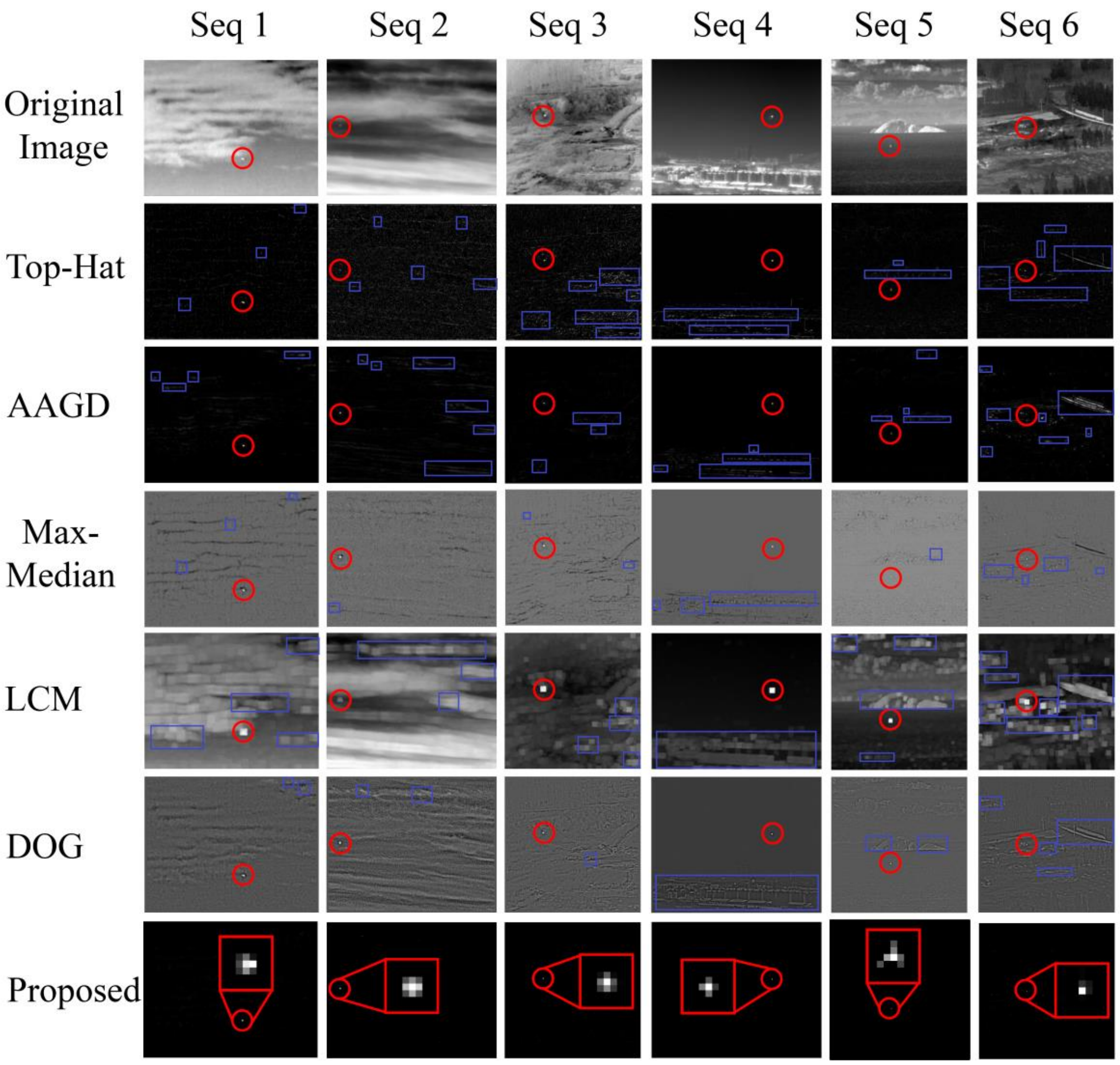

**Fig.10. The infrared image sequences used in the experiment and the processing results of each method**

Quantitative analysis was evaluating the performance of the methods through SCRG and ROC. Table 2 shows the SCRG of different methods. On sequence 1, the SCRG of the proposed method is the highest, followed by multi-scale AAGD. The SCRG of former is 10.7 times that of the latter, and 21.3 times that of the third-place Top-Hat. The proposed method has the highest SCRG on all sequences. The SCRG of Top-Hat, Max-Median, LCM and DoG are not high, not exceeding 10% of the proposed method.

Table 2 SCRG of different methods

| Sequence | Top-Hat | | AAGD | | Max-Median | | LCM | | DOG | | Proposed | |
|---|---|---|---|---|---|---|---|---|---|---|---|---|
| | $SCR_{out}$ /$SCR_{in}$ | SCRG | $SCR_{out}$ /$SCR_{in}$ | SCRG | $SCR_{out}$ /$SCR_{in}$ | SCRG | $SCR_{out}$ /$SCR_{in}$ | SCRG | $SCR_{out}$ /$SCR_{in}$ | SCRG | $SCR_{out}$ /$SCR_{in}$ | SCRG |
| Sequence 1 | 17.28 /6.92 | 2.50 | 34.46 /6.92 | 4.98 | 5.88 /6.92 | 0.85 | 10.03 /6.92 | 1.45 | 8.43 /6.92 | 1.22 | 368.69 /6.92 | **53.29** |
| Sequence 2 | 6.45 /4.02 | 1.6 | 51.40 /4.02 | 12.80 | 7.16 /4.02 | 1.78 | 6.63 /4.02 | 1.65 | 8.79 /4.02 | 2.19 | 140.02 /4.01 | **34.87** |
| Sequence 3 | 15.85 | 1.59 | 73.10 | 7.34 | 13.25 | 1.33 | 17.13 | 1.72 | 10.46 | 1.05 | 342.86 | **34.43** |
| | /9.96 | | /9.96 | | /9.96 | | /9.96 | | /9.96 | | /9.96 | |
| Sequence 4 | 54.54 /24.66 | 2.21 | 149.35 /24.66 | 6.06 | 139.10 /24.66 | 5.64 | 56.22 /24.66 | 2.28 | 24.34 /24.66 | 0.99 | 1650.8 /24.66 | **66.93** |
| Sequence 5 | 10.90 /11.83 | 0.92 | 70.54 /11.83 | 5.96 | 8.80 /11.83 | 0.74 | 6.15 /11.83 | 0.52 | 8.56 /11.83 | 0.72 | 322.86 /11.83 | **27.29** |
| Sequence 6 | 23.27 /8.03 | 2.90 | 14.48 /8.03 | 1.80 | 17.65 /8.03 | 2.20 | 6.02 /8.03 | 0.75 | 11.54 /8.03 | 1.43 | 156.34 /8.03 | **19.47** |

In the threshold segmentation stage, the higher the threshold, the lower the false alarm rate and the detection rate. At a threshold, the average detection rate and the average false alarm rate of all frames in the sequence are calculated, which is a point in the coordinate system of $Pd$-$Pf$ . Then a ROC curve can be obtained by setting the threshold from high to low. Fig.11 shows the ROC curves of different methods on six sequences. In Fig.11(a), when $Pf$ is low, the $Pd$ of the proposed method falls behind that of Max-median and DoG. But it rises to the first place when $Pf$ is more than $1\times10^{-5}$ and reaches 100% first. In Fig.11 (b), when $Pf$ is 0, $Pd$ of the proposed method is 60%, while other methods are all less than 20%. In Fig.11 (c), when $Pd$ of the proposed method reaches 100%, $Pd$ of DoG is 80%, while other methods is less than 60%. Max-median, DoG and multi-scale AAGD perform well on the whole, but on sequence 6, multi-scale AAGD has a poor effect. This is because there are a large number of bright line-shaped structures and irregular block structures in Sequence 6, while multi-scale AAGD only considers contrast and does not involve direction information, so it cannot suppress this type of clutter. The overall performance of LCM is poor, because the centroid coordinates of the connected domain are used as the target position in this paper. However, in the LCM image, the target area appears as rectangle with a similar size to the sliding window, which represents a serious diffusion phenomenon, and the center of mass of the connected domain has a drift. So the detection rate of LCM is low. The proposed method has the best performance on all sequences, of which $Pd$ first reaches 100% with the increase of $Pf$ .

To sum up, the proposed method has good capability of target enhancement and background suppression, and is effective and robust in small target detection tasks under complex background.

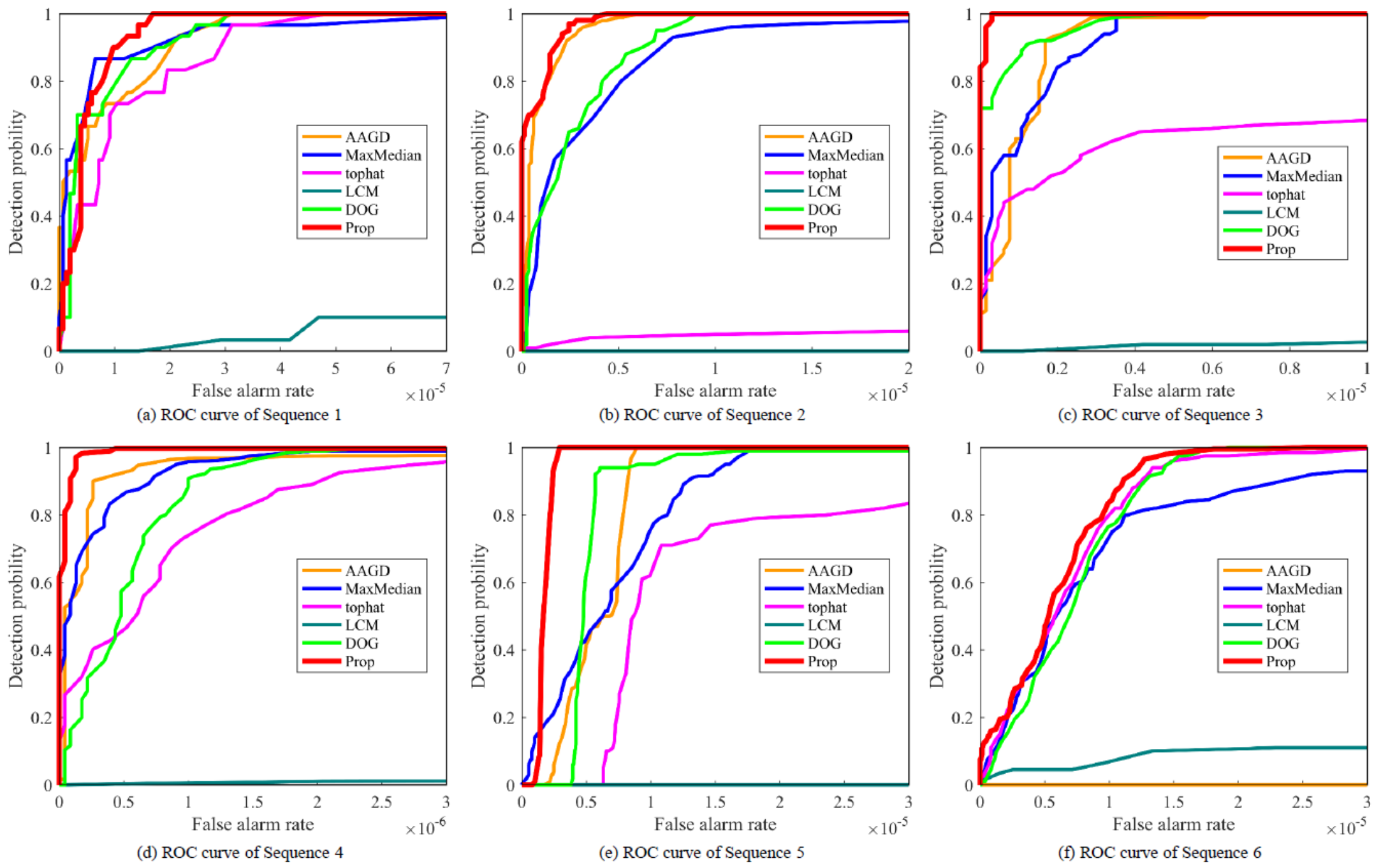

**Fig.11. ROC curves of different methods on each sequence**

## 5. Conclusion

Based on two assumptions about the features of infrared small targets, namely high contrast and isotropy, a detection method is proposed in this paper. Firstly, the salient regions with high contrast are obtained by MGD, and then the isotropy measurement of each region is obtained by calculating the Hessian matrix and its eigenvalues. In order to get the exact second partial derivative of the target, a scale estimation method is also proposed. After these two steps, the small target can be enhanced, the homogeneous background and irregular clutters are suppressed. Experiments on real image sequence suggest that the proposed method has good effectiveness and robustness, and is superior to several other common algorithms in SCRG and ROC.

## Acknowledgements

The authors would like to acknowledge National Natural Science Foundation of china (Grant No. 62001234), Postdoctoral Research Funding Program of Jiangsu Province (Grant No. 2020Z051) to provide fund for conducting experiments, China Postdoctoral Science Foundation(No.2020M681597), Natural Science Foundation of Jiangsu Province BK20200487.